\documentclass[wcp]{jmlr}

\usepackage{longtable}% for long tables

\usepackage{booktabs}
\makeatletter
\let\Ginclude@graphics\@org@Ginclude@graphics 
\makeatother

\jmlrvolume{157}
\jmlryear{2021}
\jmlrworkshop{ACML 2021}

\title[]{Robust Model-based Reinforcement Learning for Autonomous Greenhouse Control}

  \author{\Name{Wanpeng Zhang} \Email{zhangwp19@mails.tsinghua.edu.cn}\\
  \addr Tsinghua Shenzhen International Graduate School, Tsinghua University, Shenzhen, China
  \AND
  \Name{Xiaoyan Cao} \Email{holmescaoxy@gmail.com}\\
  \addr Department of Informatics, Xiamen University, Xiamen, China
  \AND
  \Name{Yao Yao} \Email{y-yao19@mails.tsinghua.edu.cn}\\
  \Name{Zhicheng An} \Email{azc19@mails.tsinghua.edu.cn}\\
  \Name{Xi Xiao} \Email{xiaox@sz.tsinghua.edu.cn}\\
  \addr Tsinghua Shenzhen International Graduate School, Tsinghua University, Shenzhen, China
  \AND
  \Name{Dijun Luo} \Email{dijunluo@tencent.com}\\
  \addr Tencent AI Lab, Shenzhen, China
  \AND
 }

\editors{Vineeth N Balasubramanian and Ivor Tsang}
\usepackage{multirow}
\usepackage{float}
\begin{document}

\maketitle

\begin{abstract}
Due to the high efficiency and less weather dependency, autonomous greenhouses provide an ideal solution to meet the increasing demand for fresh food. However, managers are faced with some challenges in finding appropriate control strategies for crop growth, since the decision space of the greenhouse control problem is an astronomical number. Therefore, an intelligent closed-loop control framework is highly desired to generate an automatic control policy. As a powerful tool for optimal control, reinforcement learning (RL) algorithms can surpass human beings' decision-making and can also be seamlessly integrated into the closed-loop control framework.
However, in complex real-world scenarios such as agricultural automation control, where the interaction with the environment is time-consuming and expensive, the application of RL algorithms encounters two main challenges, i.e., sample efficiency and safety. Although model-based RL methods can greatly mitigate the efficiency problem of greenhouse control, the safety problem has not got too much attention. In this paper, we present a model-based robust RL framework for autonomous greenhouse control to meet the sample efficiency and safety challenges. Specifically, our framework introduces an ensemble of environment models to work as a simulator and assist in policy optimization, thereby addressing the low sample efficiency problem. As for the safety concern, we propose a sample dropout module to focus more on worst-case samples, which can help improve the adaptability of the greenhouse planting policy in extreme cases. Experimental results demonstrate that our approach can learn a more effective greenhouse planting policy with better robustness than existing methods.
\end{abstract}


\section{Introduction}

The traditional agricultural production mode is highly dependent on the weather, which has been unable to meet the increasing demand for fresh and healthy food from global population growth. Around the world, some countries have gradually regarded the greenhouse industry as the main force in agricultural production~\cite{world.vegetable.map}. Modern high-tech greenhouses are equipped with standard sensors and actuators (such as heating, lighting, $CO_2$ dosing, irrigation, etc.) to empower precision agriculture. To improve crop yield and quality, managers regularly regulate a suitable environment for crop growth by overseeing the greenhouse climate and crop growth state. In addition to increasing the harvest, the corresponding energy consumption is another key consideration since natural resources are facing the challenge of exhaustion~\cite{george2018management}.
%factor which should be taken into account in the process of greenhouse control. As natural resources face the challenge of exhaustion~\cite{george2018management}, the importance of environment-friendly and resource-saving production mode is even more prominent. 
Although the automatic greenhouse is an ideal solution to deal with the food crisis, skilled managers capable of autonomous greenhouse control are scarce~\cite{greenhousegrower2018}. Furthermore, even a seasoned manager is not able to monitor and manage too many greenhouses simultaneously. 
%Meanwhile, it is also a great challenge for labors to monitor and manage multiple greenhouses simultaneously, especially when it comes to the case of greenhouses expanding in size.

To provide a favorable climate for crop growth in a modern high-tech greenhouse, growers need to manually determine control strategies, such as lighting and irrigation, according to their planting experience. Then the strategy is fed into the process computer to take effect in the greenhouse. Sensors will continuously measure the climate and crop growth state, and feedback the measured data to the grower for analysis and decision-making. The grower needs to balance production and resource consumption during a 3-5 months period~\cite{hemming2020cherry}, which implies a tremendous decision-making space. The complexity of decision-making has led to growers only giving coarse-grained control strategies, which do not make full use of the rich greenhouse states information.

With breakthroughs in AI, new technologies have been introduced into agriculture to help improve production, such as automatic pruning fruit trees~\cite{you2020efficient}, intelligent spraying systems~\cite{kim2020intelligent}, cooperative control~\cite{li2021structured}. RL has shown the potential to outperform humans in decision making via enabling both deep decision search at the macro-level and fine-grained control at the micro-level, making it well suited for automated control scenarios. By formulating autonomous greenhouse control as a Markov decision process (MDP) problem, there has been some research application of RL algorithm in an agricultural scene ~\cite{parameswaran2016arduino,wang2020deep} .
%However, these studies focus on irrigation~\cite{parameswaran2016arduino} and climate control~\cite{wang2020deep}, while more comprehensive and realistic control scene remains unstudied. In our work, we...
%When it comes to autonomous greenhouse control, it is a Markov decision process (MDP) problem, where the current control set point (such as temperature, the concentration of carbon dioxide) will influence the next crop state (such as leaf area index and fresh weight). Reinforcement learning (RL) algorithms are proven to solve the MDP problem more effectively. There are some studies focus on irrigation~\cite{parameswaran2016arduino} and climate control~\cite{wang2020deep} in an autonomous greenhouse by RL methods.

However, RL still faces two major challenges in practical application. First is the sample efficiency problem. Although RL have been achieved excellent results in some fields, they often rely on huge training samples, which is not practical in the real world. Since there is no trial-and-error cost in a virtual simulation environment, algorithms are free to perform exploratory learning and thus learn strategies that can circumvent wrong decisions. However, in the real world, where much trial-and-error implies huge costs (e.g., machine damage, crop death), it becomes a major challenge to learn decisions that can circumvent errors within limited training samples. Second, the robustness of RL is also a key challenge in real-world application scenarios. Once the environment is perturbed during the training phrase, the algorithm performance may be affected and degrade significantly. Furthermore, in the deployment phase, the inconsistency between the deployment environment and the training environment can also affect the performance of the trained strategy.

In this paper, we investigate how RL can be better applied to autonomous greenhouse control. To address the sample efficiency challenge, we introduce the model ensemble approach. In this approach, samples from the real environment are not directly handed over to the RL algorithm for policy optimization but are used to model the environment. Then the model is used to simulate the environment to add more simulated samples and accelerate the learning efficiency. To further enhance the safety of the automated planting policy, we add a sample dropout module to the RL algorithm. The module enable our algorithm to selectively discard a portion of samples with excessive reward, to focus more on worst-case samples, improving the adaptability of the planting policy in extreme cases, and solve the safety challenge of RL in a real-world deployment.

\section{Related Work}

\subsection{AI for Agriculture}

Agriculture is an area of extreme importance, which faces several challenges from sowing to harvest~\cite{bannerjee2018artificial}. With the rapid development of AI in recent years~\cite{huang2018artificial}, more and more AI technologies are being applied in agricultural automation~\cite{jha2019comprehensive}.

During crop growing, crop disease is a matter of grave concern to a farmer. Significant expertise and experience are required to detect an ailing plant and take the necessary steps for recovery. Ghaffari et al. develop an electronic nose, and intelligent systems approach to detect diseases in tomato crops~\cite{ghaffari2010early}. 
Issues on soil and irrigation management are also very vital in agriculture. Improper irrigation and soil management lead to crop loss and degraded quality. Manek and Singh compared several neural network architectures to predict rainfall using four atmospheric inputs~\cite{manek2016comparative}. Dahikar and Rode proposed a neural model to predict different crop yields using atmospheric inputs and fertilizer consumption~\cite{dahikar2014agricultural}.

Most agricultural automation works like the ones mentioned above are challenging to integrate into a holistic system. A macro-level centralized planning system for controlling entire farms remains under study. RL algorithms are often used to learn control policies in complex environments with the potential to outperform humans~\cite{schulman2015high}, as long as they have comprehensive observation information about the entire environment. An automated irrigation system is developed RL technique to control greenhouse remotely to minimize human involvement~\cite{parameswaran2016arduino}. RL algorithms are also used to optimize the autonomous greenhouse climate control to resource consumption~\cite{wang2020deep}.

Also, some works on agriculture with robotics as well, where mobile robots are internally powered by some AI algorithms to facilitate their jobs. In our opinion, this direction of research is promising in the sense that, conceivably, autonomous robots can significantly boost the management of large-scale crop farms as human labor becomes increasingly expensive and conventional machines are not intelligent enough. A few works down the line are~\cite{2002Online} and~\cite{zhou2014vision} where they discuss online learning of robots and visual navigation of mobile robots.

% However, there have been new works in recent years that focus on this challenge, looking at how to integrate reinforcement learning into greenhouse control systems. Fanyu et al.~\cite{bu2019smart} present a review of combining multiple reinforcement learning algorithms with IoT devices, which serves as an essential guide for our work's overall framework. Andrey et al.~\cite{somov2018pervasive} introduced model-based reinforcement learning for overall control of greenhouses through a central control policy. However, they did not focus on addressing the security concerns of reinforcement learning in agriculture due to its drawbacks of poor security in practical application scenarios.

\subsection{Challenges of RL in Agricultural Applications}

In recent years, RL has achieved remarkable results in a wide range of areas, including simulated control problems \cite{levine2016end}, outperforming human performances on Go and games \cite{mnih2015human,silver2016mastering}. However, applying reinforcement learning to agricultural applications requires addressing the two major challenges of sample efficiency and robustness.

% However, RL algorithms face two major challenges, sample efficiency and robustness, making it hard to solve real-world problems.

% The main challenge regarding sample efficiency is that reinforcement learning relies on a large number of training samples to learn a useful policy, and to address this challenge, some approaches incorporating simulation techniques have been proposed. 

Model-based methods have shown excellent abilities to reduce the sample complexity \cite{deisenroth2013survey}. Previous works \cite{levine2016end,Chua2018DeepModels,janner2019trust} have empirically shown significant sample efficiency improvements. However, model accuracy is a major barrier for policy quality, and it is challenging to build an accurate model in high-dimensional tasks \cite{abbeel2006using}. Thus the policy learned on inaccurate models typically leads to performance degradation due to cumulative model error \cite{sutton1996model,asadi2019combating}. While improving sample efficiency, the control policy is affected by the discrepancy between the simulator and the real environment. For this problem, previous works (e.g. PETS  \cite{Chua2018DeepModels}, ME-TRPO \cite{kurutach2018model}, SLBO \cite{luo2018algorithmic}, MB-MPO \cite{clavera2018model})  use ensembles of bootstrapped probabilistic transition models to properly incorporate two kinds of uncertainty into the transition model. Concretely, individual probabilistic models capture aleatoric uncertainty or the noise due to the inherent stochasticity. The bootstrapping ensemble procedure can capture epistemic uncertainty or uncertainty in the model parameters aroused from insufficient training data. 
Empirical works \cite{levine2016end,janner2019trust,yao2021sample} have demonstrated that the ensemble of probabilistic models is an efficient way to handle both of two uncertainties, allowing for a competitive model-based learning algorithm.

\begin{figure}[t]
\centering
\includegraphics[width=0.81\textwidth]{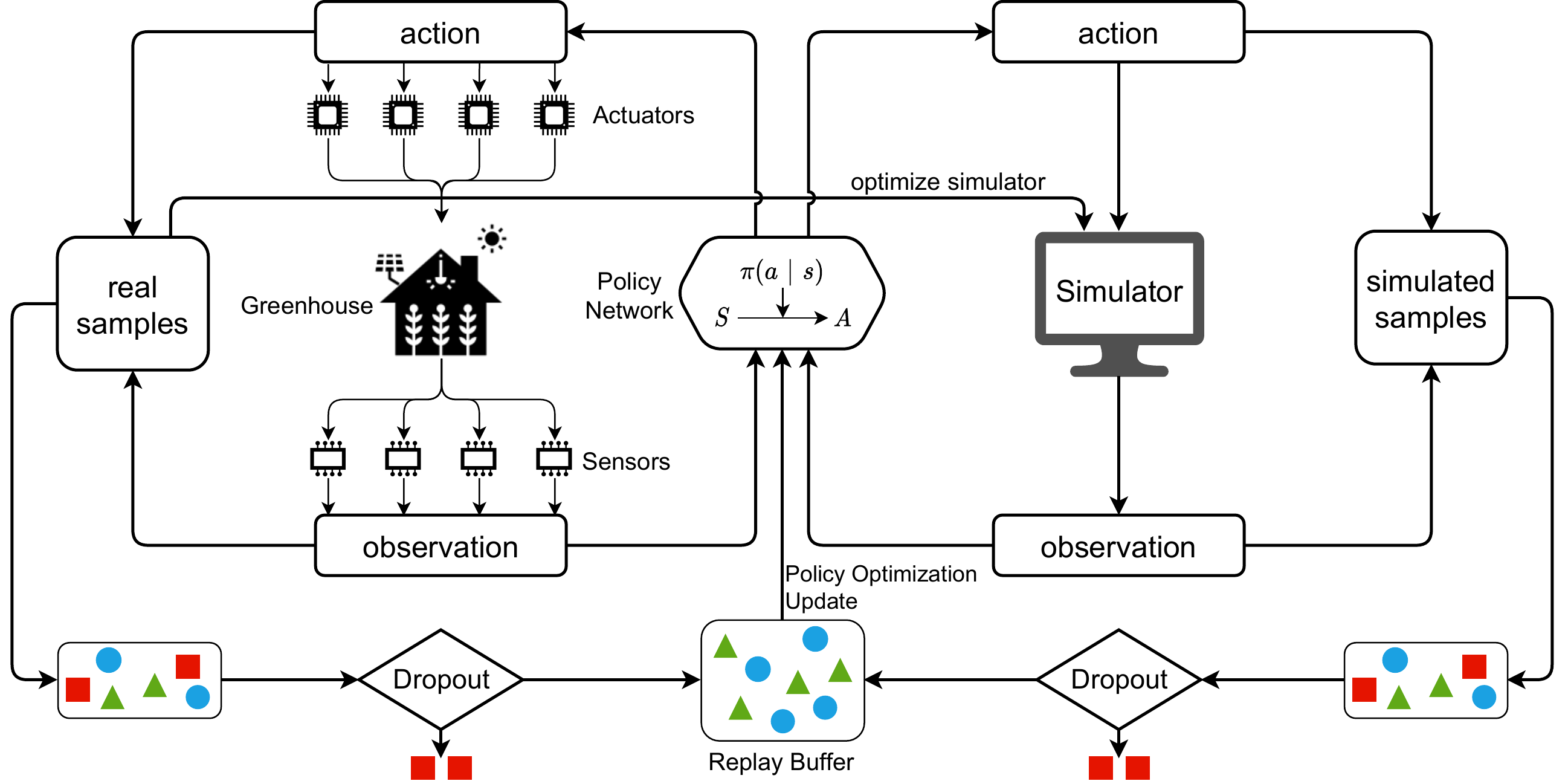}
\caption{A modular framework for greenhouse automation. It mainly consists of a greenhouse environment, a crop growth simulator and the planting algorithm. The core part is the policy module, which receives observations to decide the next action. The RL algorithm uses samples to optimize policies continuously. These three parts alternate in a cycle that eventually leads to an automated planting policy.}
\label{fig:framework}
\end{figure}

Moreover, the robustness issue of RL is one of the challenges hindering its application to complex tasks. Though current state-of-the-art model-based methods have achieved outstanding performance, the derived policies are often only adequate for the environment in which they are trained, and when deployed to perturbed real environments, the policy performance tends to degrade dramatically or even behave dangerously. Robust control \cite{zhou1998essentials} is a branch of control theory focused on finding optimal policy under worst-case situations. Policy learned through robust control has better generalization performance. Robust adversarial RL methods, such as the RARL \cite{pinto2017robust} and NR-MDP \cite{tessler2019action} methods, learn robust decision policy by adversarially adding minimax objects, while EPOpt method introduces conditional value at risk (CVaR) \cite{Tamar2015OptimizingSampling,rajeswaran2016epopt}, by optimizing the CVaR object, a robust policy with better performance is obtained. We follow these ideas and design a sample dropout module that improves safety in tomato greenhouse automation.

\section{Notations and Preliminaries}

% \subsection{Reinforcement Learning}

We consider a Markov decision process (MDP), defined by $(\mathcal{S},\mathcal{A}, \mathcal{P}, r, \gamma)$ where $\mathcal{S}\in\mathbb{R}^{d_s}$ is the state space, $\mathcal{A}\in\mathbb{R}^{d_a}$ is the action space, $r(s,a):\mathcal{S}\times\mathcal{A}\mapsto\mathbb{R}$ is the reward function, $\gamma\in [0,1]$ is the discount factor, and $\mathcal{P}(s'|s,a):\mathcal{S}\times\mathcal{A}\times\mathcal{S}\mapsto[0,1]$ is the probability distribution of the next state given current state $s$ and action $a$, or use the form $s'=\mathcal{P}(s,a) : \mathcal{S}\times\mathcal{A}\mapsto\mathcal{S}$ denotes the state transition function when the environment is deterministic.

Let $V^{\pi_\theta,\mathcal{P}}(s)$ denotes the expected return or expectation of cumulative rewards starting from initial state $s$, i.e., the expected sum of discounted rewards following policy $\pi_\theta(a|s)$ and state transition function $\mathcal{P}(s,a)$:
\begin{equation}\label{def:eta-s}
    V^{\pi_\theta,\mathcal{P}}(s) = \underset{\{a_0,s_1,\ldots\} \sim \pi_\theta,\mathcal{P}}{\mathbb{E}}\left[\sum_{t=0}^\infty\gamma^t r(s_t,a_t)\mid s_0=s\right]
\end{equation}
For simplicity of symbol, let $V^{\pi_\theta,\mathcal{P}}$ denotes the expected return over random initial states: $    V^{\pi_\theta,\mathcal{P}} = \underset{s_0\in\mathcal{S}}{\mathbb{E}} \left[V^{\pi_\theta,\mathcal{P}}(s_0)\right]$. The goal of RL is to maximize the expected return by finding the optimal decision policy
\begin{equation}
    \pi_\theta^* = {\arg\max}_{\pi_\theta}\ V^{{\pi_\theta},\mathcal{P}}
\end{equation}

In model-based RL, an approximated transition model $\mathcal{M}(s, a)$ is learned by interacting with the environment, the policy $\pi(a|s)$ is then optimized using the model-free method with samples from the environment and data generated by the model. We use the parametric notation $\mathcal{M}_\phi, \phi\in\Phi$ to specifically denote the model trained by a neural network, where $\Phi$ is the parameter space of models.

\section{Proposed Approach}

\subsection{System overview}

We propose the framework shown in the Fig. \ref{fig:framework}, which mainly consists of three components that can be executed asynchronously:

\begin{enumerate}
    \item The collection of real samples through a real agricultural greenhouse environment and a large number of IoT hardware devices.
    \item The generation of samples through rapid simulation in a tomato growth simulator.
    \item Based on the observation samples, the RL algorithm is used to continually train and optimize the policy, which makes decisions based on current information and performs the next action to obtain new observations, leading to the next cycle.
\end{enumerate}

% \begin{figure}
% 	\centering
% 	\subfloat[temperature \& humidity sensor]{\includegraphics[width=0.27\linewidth]{imgs/TempHumidity-sensor.jpg}}\hfil
% 	\subfloat[CO$_2$ sensor]{\includegraphics[width=0.27\linewidth]{imgs/CO2-sensor.jpg}}\hfil
% 	\subfloat[PAR sensor]{\includegraphics[width=0.27\linewidth]{imgs/PAR-sensor.jpg}}\\
%     \subfloat[ventilation controller]{\includegraphics[width=0.27\linewidth]{imgs/Ventilation-controller.jpg}}\hfil
% 	\subfloat[CO$_2$ producer]{\includegraphics[width=0.27\linewidth]{imgs/CO2-producer.jpg}}\hfil
% 	\subfloat[fertigation controller]{\includegraphics[width=0.27\linewidth]{imgs/Fertigation-controller.jpg}}\\	
% 	\caption{Monitoring sensors and control equipment in our greenhouse.}
% \end{figure}

\begin{figure}[t]
	\centering
	\subfigure[]{\includegraphics[width=0.15\linewidth]{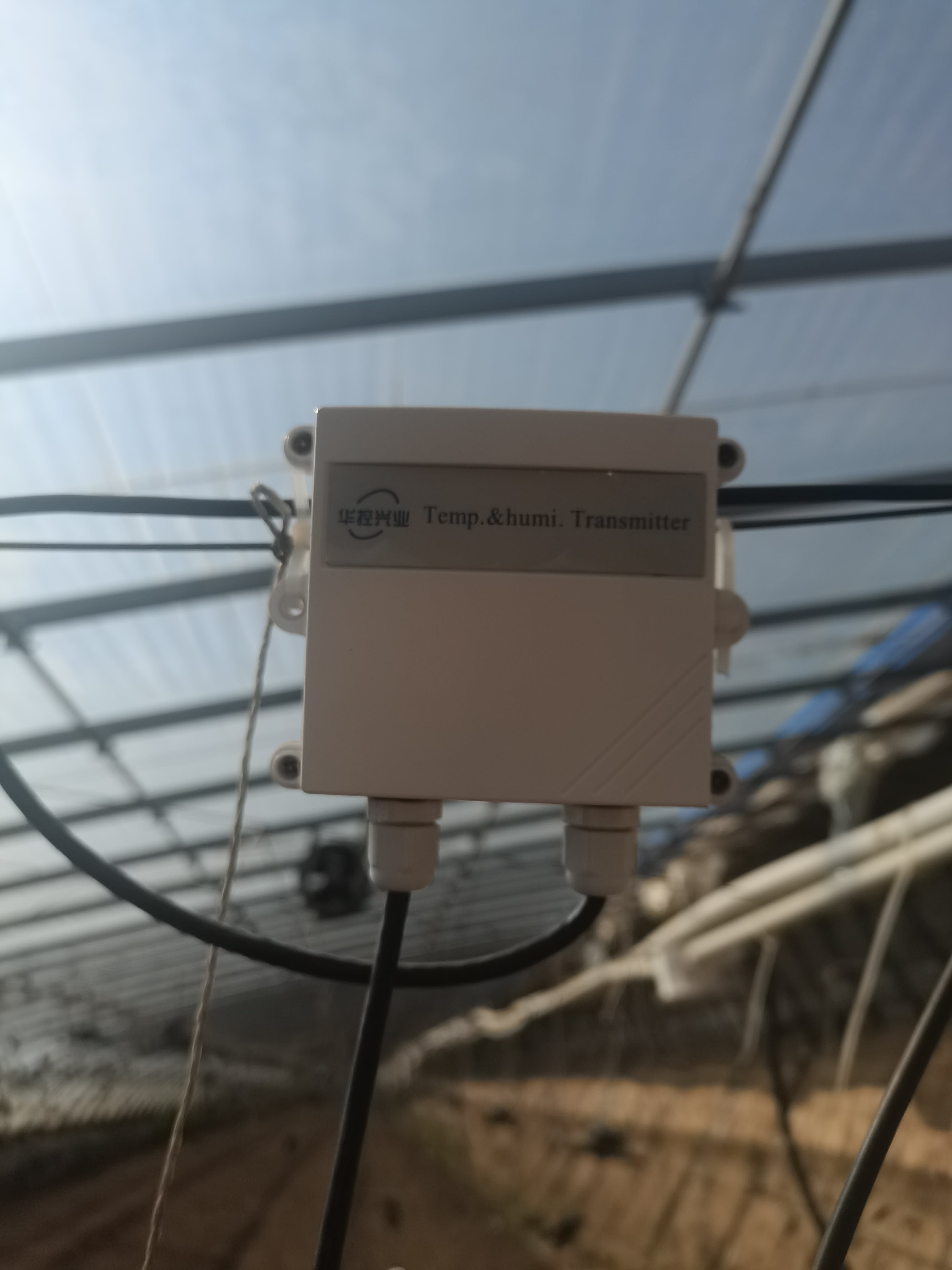}}\hfil
	\subfigure[]{\includegraphics[width=0.15\linewidth]{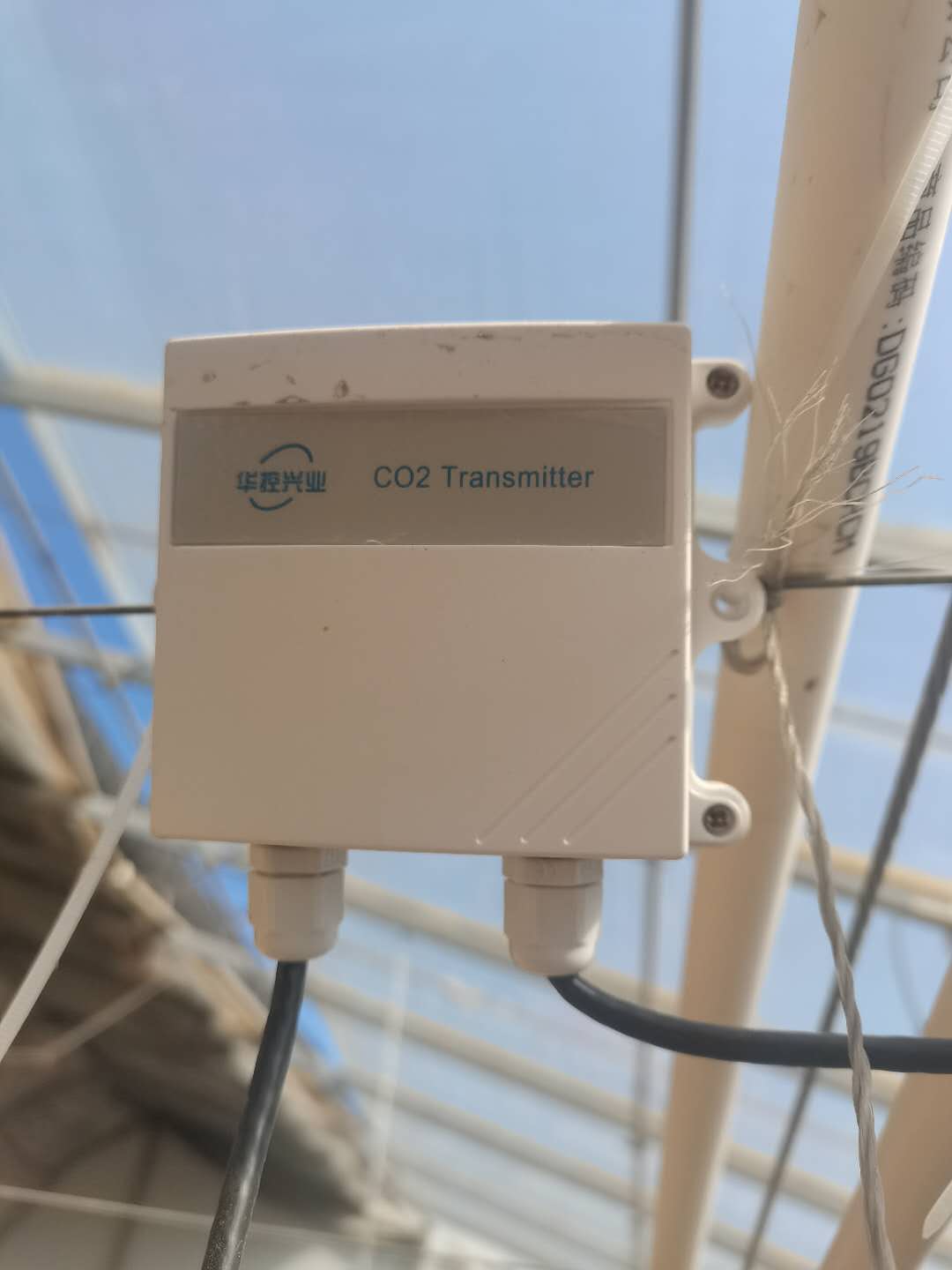}}\hfil
	\subfigure[]{\includegraphics[width=0.15\linewidth]{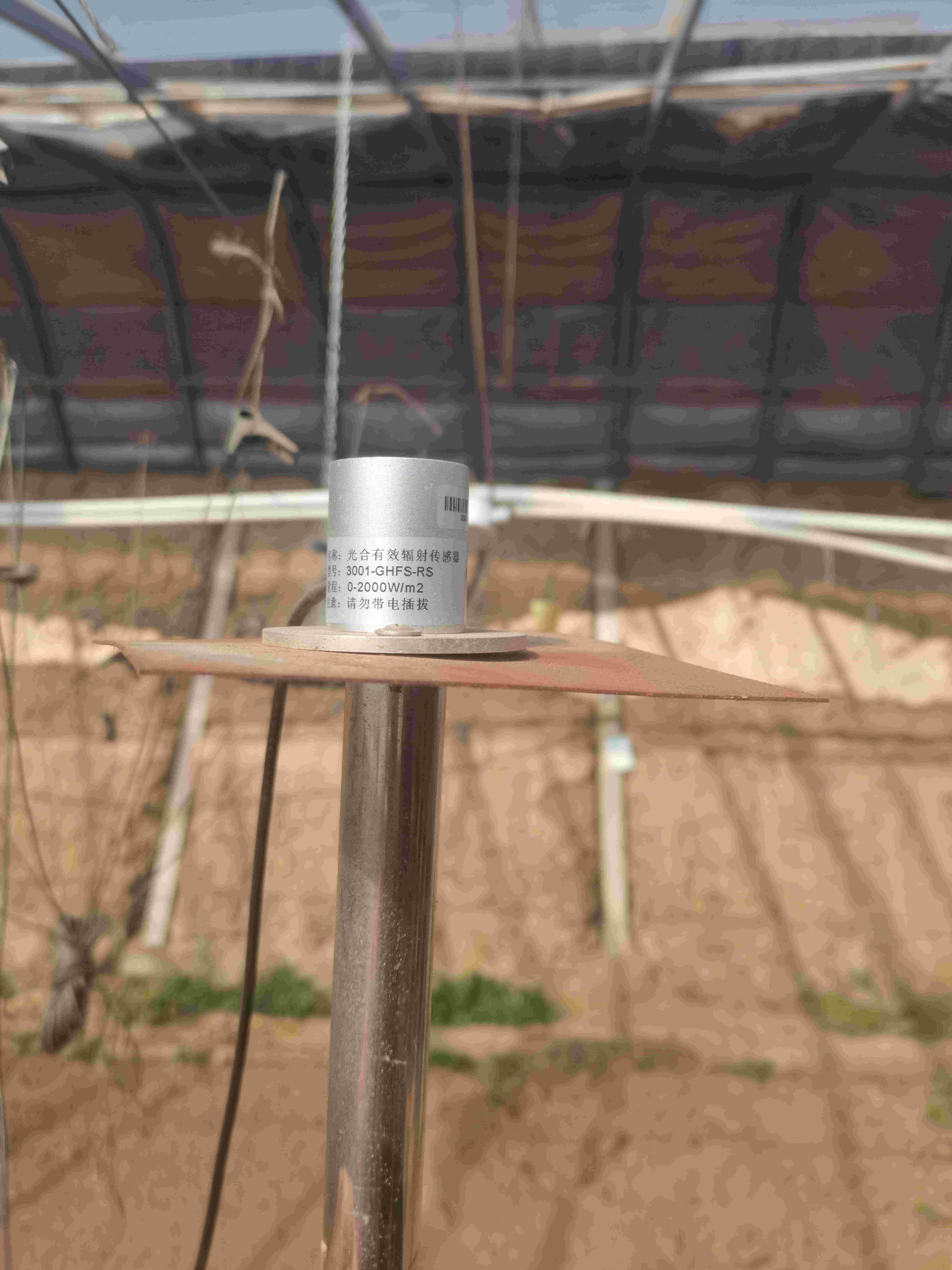}}\hfil
    \subfigure[]{\includegraphics[width=0.15\linewidth]{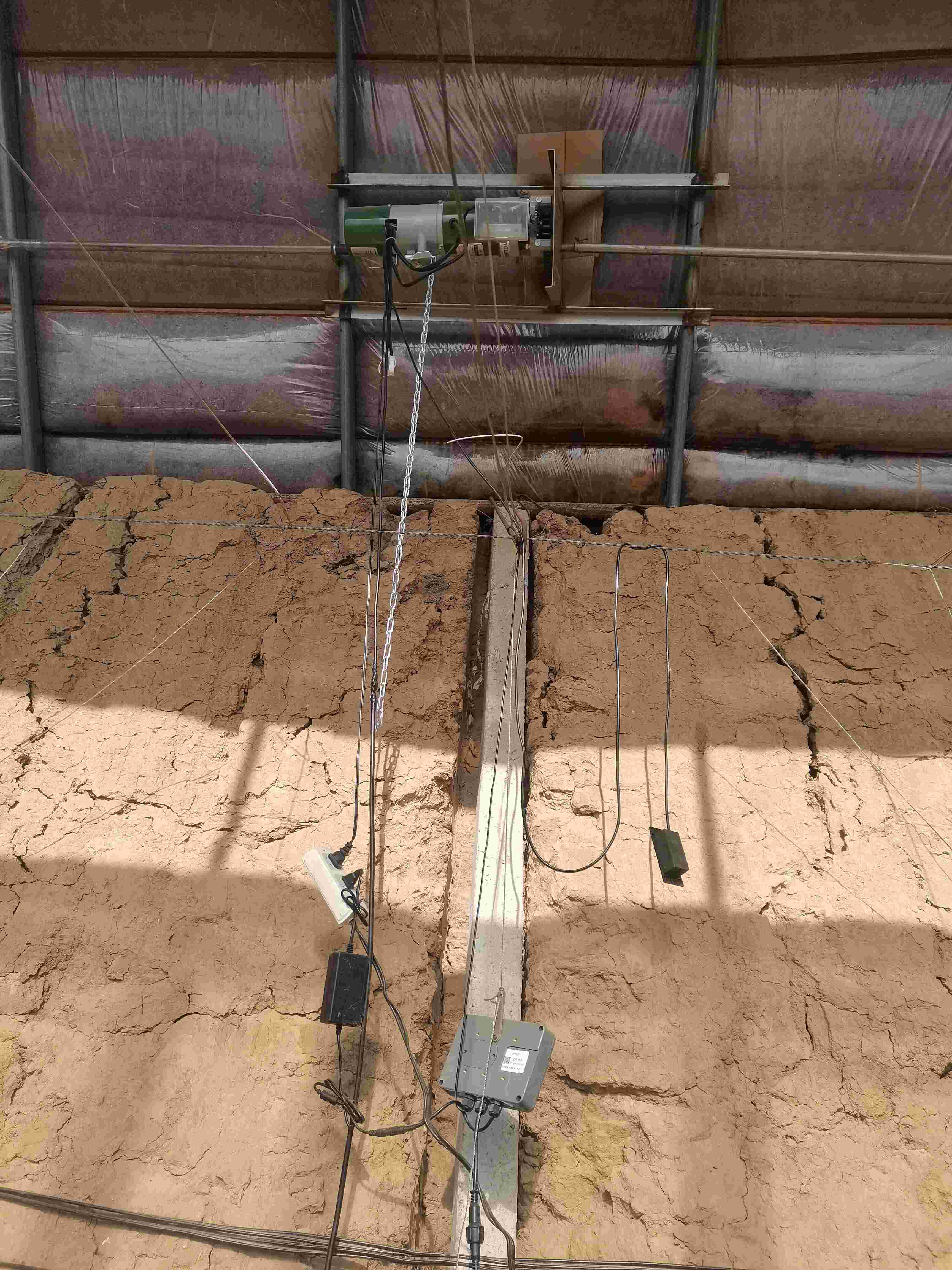}}\hfil
	\subfigure[]{\includegraphics[width=0.15\linewidth]{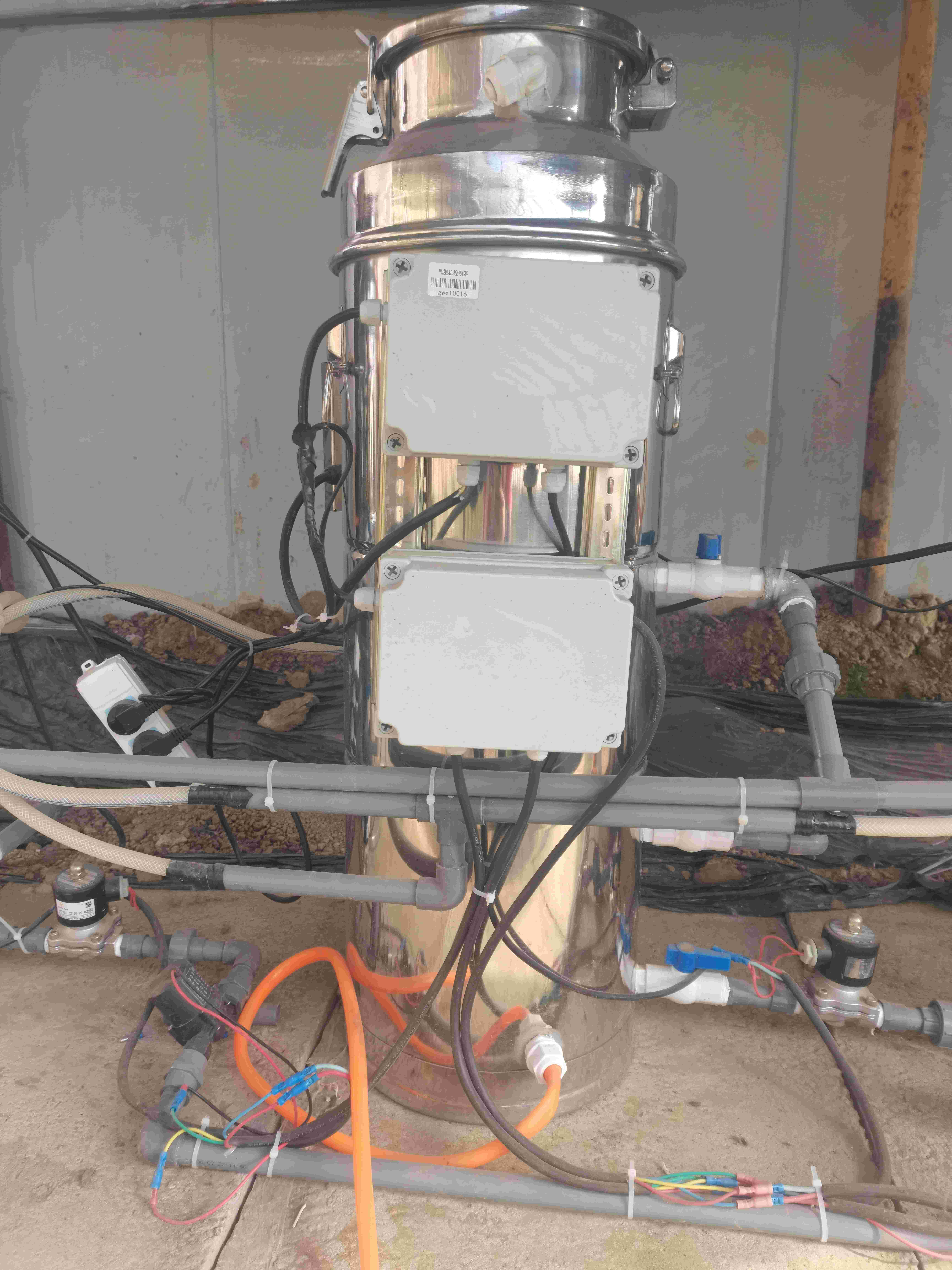}}\hfil
	\subfigure[]{\includegraphics[width=0.15\linewidth]{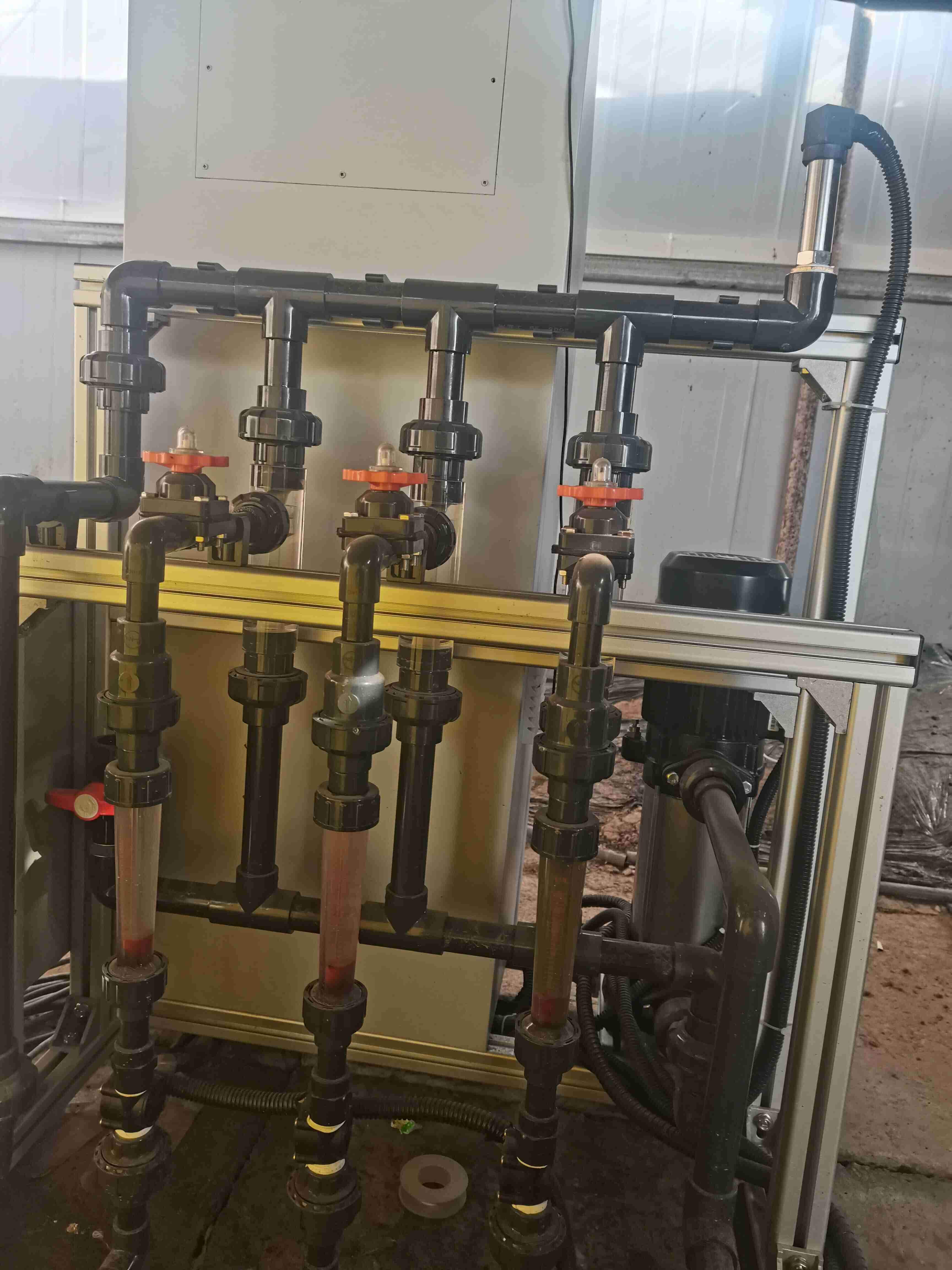}}\\	
	\caption{Equipments in our greenhouse. (a)temperature sensor. (b)CO$_2$ sensor. (c)PAR sensor. (d)ventilation controller. (e)CO$_2$ producer. (f)fertigation controller.}
\end{figure}

Specifically, we use multiple sensors to obtain observations, such as temperature, humidity, CO$_2$ concentration, etc. Here are the details of how the main sensors work:

\begin{itemize}
    \item Temperature, humidity and CO$_2$ sensor: multiple measuring boxes are hung in the greenhouse and their readings are averaged (sometimes with different weighting factors for different locations) to get the parameters of temperature, humidity, and CO$_2$ as the feed for heating, humidity, ventilation, and CO$_2$ supply.
    \item PAR sensor: the greenhouse keeps track of the supply of PAR parameter by measuring the Photosynthetically Active Radiation. In the simulated greenhouse, a PAR sensor is modeled to describe the PAR-level just above the crop.
\end{itemize}

The observation samples are then transmitted to the decision-making policy module, which can generate corresponding actions. These actions are executed in the greenhouse through multiple actuators. Here are the details of how the main actuators work:

\begin{itemize}
    \item Ventilation controller: when the greenhouse temperature exceeds the ventilation setpoint, the vents will be opened to a certain extent. The vents' opening angle is expressed as a percentage, and it is determined by the deviation between the current greenhouse temperature and the temperature-setpoint.
    \item CO$_2$ producer: when the CO$_2$ concentration drops below the setpoint, the CO$_2$ producer will supply CO$_2$ through a piping network.
    \item Fertigation controller: The crops will be automatically irrigated by drip irrigation according to the related setpoints, including irrigation time and watering quantity.
\end{itemize}

All the observation and action samples are stored in the data cache to optimize both the simulator and policy.

\subsection{Algorithm}

Although RL is a powerful tool for policy learning, it faces a significant challenge in training efficiency, often relying on a large number of interactions with the environment. Based on the data collected, we developed a tomato growth simulator close to the real environment. The simulator allows us to rapidly perform action interactions and generate simulation samples that simulate a tomato crop's entire growth cycle within seconds. These samples are then provided to RL algorithms for policy optimization.

\begin{algorithm}[t]
\caption{Robust Model-based RL for Autonomous Greenhouse Control}
\DontPrintSemicolon
\SetKwFunction{FDrop}{Dropout}
\SetKwProg{Fn}{Function}{:}{}
\Fn{\FDrop{$\mathcal{B}$, $p$}}{
    Calculate $r_p(\mathcal{B})$: the $p$-percentile of batch $\mathcal{B}$\\
    \For{$x\in \mathcal{B}$}{
        \If{$r(x)\leq r_p(\mathcal{B})$}{
            Fill $x$ into $\mathcal{B}_{\mathrm{drop}}$
        }
    }
    \KwRet $\mathcal{B}_{\mathrm{drop}}$
}
\SetKwFunction{FMain}{Main}
\SetKwProg{Fn}{Function}{:}{}
\Fn{\FMain}{
Initialize hyperparameters, policy $\pi_\theta$, environment replay buffer $\mathcal{D}_{\mathrm{real}}$, simulator replay buffer $\mathcal{D}_{\mathrm{sim}}$\\
\For{$N_\mathrm{epoch}$ iterations}{
    Take an action in the greenhouse environment using policy $\pi_\theta$; add samples to $\mathcal{D}_{\mathrm{real}}$\\
    Mask $\mathcal{D}_{\mathrm{real}}$ into $\left\{\mathcal{D}_{\mathrm{1}}, \mathcal{D}_{\mathrm{2}}, \ldots, \mathcal{D}_{\mathrm{N}}\right\}$\\
    \For{$N_\mathrm{train}$ iterations}{
        Load the pre-trained model and train on $\left\{\mathcal{D}_{\mathrm{1}}, \mathcal{D}_{\mathrm{2}}, \ldots, \mathcal{D}_{\mathrm{N}}\right\}$\\
        Get the trained model ensemble collection $\mathcal{M} = \{\mathcal{M}_{\phi_1},\mathcal{M}_{\phi_2},\ldots,\mathcal{M}_{\phi_{N}}\}$ with clip restriction\\
        \For{$t=1,2,\ldots ,T$}{
            Select a model $\mathcal{M}_t$ from $\mathcal{M}$ with probability $\mathrm{Pr}\{\mathcal{M}_t=\mathcal{M}_{\phi_i}\mid i\sim P_{\mathcal{M}}, i\in\{1,2,\ldots,N\}\}$\\
            Perform rollouts on model $\mathcal{M}_t$ with policy $\pi_\theta$ and get samples $x=\left(s_{t+1},s_t,a_t\right)$ \\
            Fill these samples into batch $\mathcal{B}^{\pi_\theta,\mathcal{M}}$\\
        }
        $\mathcal{B}^{\pi_\theta,\mathcal{M}}_p$ = \FDrop{$\mathcal{B}^{\pi_\theta,\mathcal{M}}$, $p$}; Fill the data of $\mathcal{B}^{\pi,\mathcal{M}^\alpha}_p$ into $\mathcal{D}_\mathrm{sim}$
    }
    $\mathcal{B}^{\pi_\theta}_p$ = \FDrop{$\mathcal{D}_{\mathrm{real}}$, $p$}; $\mathcal{D}_{\mathrm{real}} = \mathcal{B}^{\pi_\theta}_p$\\
    Optimize policy $\pi_\theta$ on $\mathcal{D}_{\mathrm{sim}}$ and $\mathcal{D}_{\mathrm{real}}$: $\theta\leftarrow \theta - \lambda\nabla_\theta V^{{\pi_\theta},\mathcal{M}}_p(\mathcal{D}_{\mathrm{sim}}; \mathcal{D}_{\mathrm{real}})$
}
\KwRet
}
\label{algo:our-method}
\end{algorithm}

Specifically, we firstly pre-train a model based on the cumulative greenhouse crop data, i.e., $\mathcal{D}_{\text{real}}$, to express the relationship between crop growth and the parameters of greenhouse. The initialization of model parameters is based on expert knowledge, which can effectively alleviate the sample consumption in the early stage of model learning. Besides, we add restrictions to the simulation states based on prior knowledge in agriculture to avoid unreasonable anomalies (e.g., excessive temperature and CO$_2$ concentration) due to model generalization, enabling the simulator to simulate more realistic and complex situations and improve the robustness of the learned planting policy, thus improving the adaptability of the policies to a variety of complex scenarios. And with the idea of Dyna-style RL~(\cite{Sutton1991DynaReacting}), we continually transmit the collected real samples to the simulator for further optimization of the tomato growth model in the policy optimization cycle, which enables the simulator to provide simulation samples for the framework more efficiently and accurately. We also use the model ensemble approach in order to capture the uncertainty in the real greenhouse environment.

We first build the binary masks $h_{i,j}$ from the Bernoulli distribution with parameter $p\in(0, 1]$, and we perform the bootstrap mask
\begin{equation}
    H_{j}=\left\{h_{i,j} \sim \text { Bernoulli }(p) \mid i \in\{1, \ldots, N\}\right\}
\end{equation}
on each sample data $d_j\in\mathcal{D}_{\text{real}}$, then we can generate $N$ subsets: $ \mathcal{D}_i=\left\{h_{i,j}\bigodot d_j\mid d_j\in\mathcal{D}_{\text{real}}\right\} $ where $\bigodot$ indicates that $d_j$ is retained in the set $\mathcal{D}_i$ if $h_{i,j}\neq 0$, otherwise not. And then, based on the pre-trained simulator, we use the new data from the subsets to fine-tune the model. We learn a collection of fine-tuned simulator models $\mathcal{M} \doteq \{\mathcal{M}_{\phi_1},\mathcal{M}_{\phi_2},\ldots,\mathcal{M}_{\phi_N}\}$. We use parametric notation $\mathcal{M}_\phi, \phi\in\Phi$ to specifically denote the model trained by neural network, where $\Phi$ is the parameter space of models. Each member of the collection $\mathcal{M}$ is a probabilistic neural network whose outputs $\mu_{\phi_i},\sigma_{\phi_i}$ parametrize a Guassian distribution:
\begin{equation}
    s^\prime = \mathcal{M}_{\phi_i}(s,a) \sim \mathcal{N}(\mu_{\phi_i}(s,a),\sigma_{\phi_i}(s,a))
\end{equation}
The corresponding loss of the simulator models is
\begin{equation}
    \mathcal{L}_{\mathcal{M}_{\phi_i}} =  \sum_{t=1}^N[\mu_{\phi_i}(s_t,a_t)-s_{t+1}]^\top\sigma_{\phi_i}^{-1}(s_t,a_t)[\mu_{\phi_i}(s_t,a_t)-s_{t+1}] + \log|\sigma_{\phi_i}(s_t,a_t)|
\end{equation}

% \begin{figure}
% \centering
% \includegraphics[width=0.3\linewidth]{imgs/simulator.pdf}
% \caption{We pretrain a tomato growth model using real data, and the samples collected during the training process will be used for further optimization of the model.}
% \label{fig:simulator}
% \end{figure}

In our method, the simulator is learned to replace the real environment $\mathcal{P}$. The policy ${\pi_\theta}(a|s)$ is then optimized using the samples from both the environment and simulator. Optimizing the expected return in a general way as RL methods allows us to learn a policy that performs best in expectation over the simulator. However, best expectation doesn't mean that the result policies can perform well in each individual model, since there could be high variability in performance for different models. This instability typically leads to risky decisions when facing poorly-informed states at deployment.

Inspired by previous works \cite{Tamar2015OptimizingSampling,rajeswaran2016epopt} which optimize conditional value at risk (CVaR) to explicitly seek a robust policy, we add a sample dropout module to the RL algorithm, which selectively discards a portion of samples with excessive reward, to focus more on worst-case samples, improving the adaptability of the planting policy in extreme situations, and solve the safety challenges of RL in real-world deployment, aiming to further enhance the safety of the automated planting policy. 

More specifically, to generate a prediction from our model collection, we first select a model $\mathcal{M}_t$ with probability
\begin{equation}
    \mathrm{Pr}\{\mathcal{M}_t=\mathcal{M}_{\phi_i}\mid i\sim P_{\mathcal{M}}, i\in\{1,2,\ldots,N\}\}
\end{equation}
at each time step $t$, where $P_\mathcal{M}$ is a probability distribution (defaults to a random distribution). We then use the model $\mathcal{M}_t$ to interact with the policy ${\pi_\theta}$ and perform a simulated rollout using the selected model, i.e., $s_{t+1}\sim \mathcal{M}_t(s_t,a_t), a_{t+1}\sim {\pi_\theta}(a|s_t)$. In this way, we can quickly generate a large number of crop growth simulation samples, and our simulator can be improved as we receive real samples from the greenhouse during the RL cycle. Then we fill these rollout samples $x=\left(s_{t+1},s_t,a_t\right)$ into a batch and retain a $p$-percentile dropout subset with more pessimistic rewards. We use $\mathcal{B}_p^{{\pi_\theta},\mathcal{M}}$ to denote the $p$-percentile dropout rollout batch:
\begin{equation}\label{def:batch-alpha-beta}
    \mathcal{B}_p^{{\pi_\theta},\mathcal{M}}=\left\{x|x\in\mathcal{B}^{{\pi_\theta},\mathcal{M}},r(x)\leq r_p(\mathcal{B}^{{\pi_\theta},\mathcal{M}})\right\}
\end{equation}
where 
\begin{equation}
    \mathcal{B}^{{\pi_\theta},\mathcal{M}}=\left\{x|x\triangleq\left(s_{t+1},s_t,a_t\right)\sim{\pi_\theta},\mathcal{M}\right\}
\end{equation}
is the sample batch collected by performing policy $\pi_\theta$ on the model ensemble $\mathcal{M}$ and $r_p(\mathcal{B}^{{\pi_\theta},\mathcal{M}})$ is $p$-percentile of reward values in batch $\mathcal{B}^{{\pi_\theta},\mathcal{M}}$. The expected return of dropout batch rollouts is denoted by $V^{{\pi_\theta},\mathcal{M}}_p$:

\begin{equation}\label{def:eta-beta}
    V^{{\pi_\theta},\mathcal{M}}_p=\mathbb{E}\left[{\sum}_{\{s_0,a_0,\ldots\} \sim\mathcal{B}_p^{{\pi_\theta},\mathcal{M}}}\left[\gamma^t r(s_t,a_t)\right]\right]
\end{equation}

Then, we can perform policy gradient update by

\begin{equation}
    \boldsymbol{\theta}^\prime=\boldsymbol{\theta}-\lambda \nabla_{\boldsymbol{\theta}} V^{{\pi_\theta},\mathcal{M}}_p
\end{equation}
Where $\lambda$ is the learning rate. The overall pseudo code is shown as Algorithm \ref{algo:our-method}.

\section{Experiments}

With the sensors and actuators deployed in our tomato greenhouse, we collect 22 kinds of observation variables, constituting a 275-dimensional observation space $\mathcal{S}$, and 6 kinds of control variables, constituting a 52-dimensional action space $\mathcal{A}$. Also, we use the $Net profit$ (USD/m$^2$) as the target reward for training. $Net profit = Gains-Costs$, where $Gains$ are obtained through yields and price, and $Costs$ include resource consumption (electricity, heat, $CO_2$, and water) and crop maintenance costs. The details of observation space and action space are shown in the Table \ref{tab:obs-sapce}. Since it’s difficult to have a ground truth simulator in a complex environment like the greenhouse. However, due to the long experiment period in real world, we have to consider using the simulator as the test environment (different from our pre-trained models). We use a commercial simulator designed by human experts, which simulates the real greenhouse environment as much as possible, but it can only be used for testing due to low computational efficiency. Since the simulator is a black-box model independent of the training environment, it’s reasonable and fair for testing.

\begin{table}[ht]
\centering
\caption{Observation Space (Above) and Action Sapce (Below)}
\label{tab:obs-sapce}
\begin{tabular}{cccc}
\toprule[1pt]
\textbf{Name}                                     & \textbf{Min} & \textbf{Max} & \textbf{Dim} \\
\midrule[1pt]
temperature setpoint                              & 13           & 32           & 24           \\
CO$_2$ setpoint                                   & 400          & 1000         & 24           \\
light-on time                                     & 0            & 24           & 1            \\
light-off time                                    & 0            & 24           & 1            \\
irrigation start time                             & 0            & 24           & 1            \\
irrigation stop time                              & 0            & 24           & 1            \\
outside solar radiation                           & 0            & 2000         & 24           \\
outside temperature                               & -30          & 50           & 24           \\
outside humidity                                  & 0            & 100          & 24           \\
wind speed                                        & 0            & 25           & 24           \\
virtual sky temperature                           & -20          & 20           & 24           \\
greenhouse air temperature                        & -30          & 100          & 24           \\
greenhouse air humidity                            & 0            & 100          & 24           \\
greenhouse air CO$_2$ concentration               & 400          & 1000         & 24           \\
light intensity just above crop                   & 0            & 2000         & 24           \\
cumulative amount of irrigation per day           & 0            & 10           & 1            \\
cumulative amount of drain per day                & 0            & 10           & 1            \\
leaf area index                                   & 0            & 10           & 1            \\
current number of growing fruits                  & 0            & 1000         & 1            \\
cumulative harvest in terms of fruit fresh weight & 0            & 100          & 1            \\
cumulative harvest in terms of fruit dry weight   & 0            & 100          & 1            \\
planting days                                     & 0            & 365          & 1            \\
\midrule[1pt]
temperature setpoint          & 13           & 32           & 24           \\
CO$_2$ setpoint               & 400          & 1000         & 24           \\
light on time                 & 0            & 24           & 1            \\
light off time                & 0            & 24           & 1            \\
irrigation start time         & 0            & 24           & 1            \\
irrigation stop time          & 0            & 24           & 1            \\
\bottomrule[1pt]
\end{tabular}
\end{table}

% \begin{table}[ht]
% \centering
% \caption{Action Space}
% \label{tab:action-space}
% \begin{tabular}{cccc}
% \toprule[1pt]
% \textbf{Name}                 & \textbf{Min} & \textbf{Max} & \textbf{Dim} \\
% \midrule[1pt]
% temperature setpoint          & 13           & 32           & 24           \\
% CO$_2$ setpoint               & 400          & 1000         & 24           \\
% light on time                 & 0            & 24           & 1            \\
% light off time                & 0            & 24           & 1            \\
% irrigation start time         & 0            & 24           & 1            \\
% irrigation stop time          & 0            & 24           & 1            \\
% \bottomrule[1pt]
% \end{tabular}
% \end{table}

\subsection{Analysis of Performance}

\begin{figure}
\centering
\includegraphics[width=0.7\textwidth]{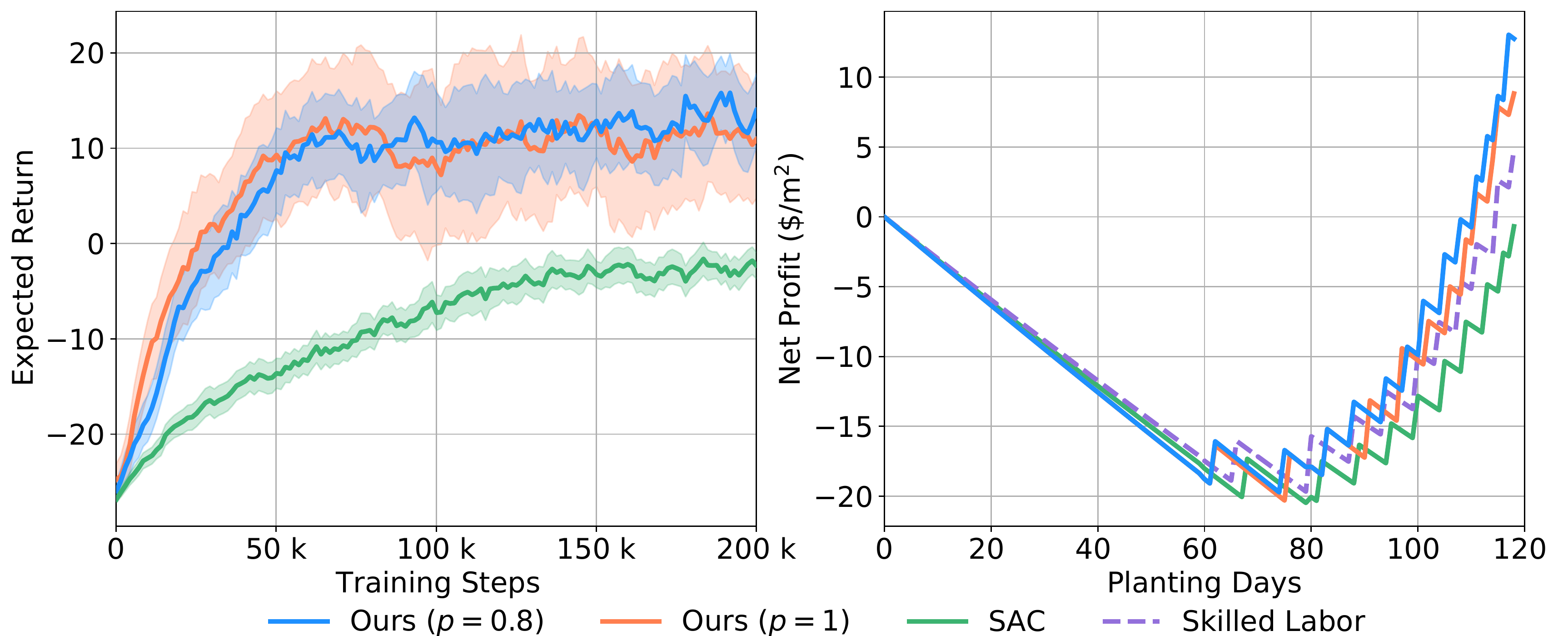}
\caption{(1) The left plot shows the training curves for our method with dropout ($p=0.8$), our method without dropout ($p=1$) and the SAC baseline (the original version). Each with 5 seeds for training. Solid curves indicate the mean of all trials with different seeds. Shaded regions correspond to standard deviation among trials; (2) The right plot shows the evaluation curves of the policies trained by all the algorithms. The purple dashed line is the reference profit for skilled labor.}
\label{fig:results}
\end{figure}

We train two versions of our method on the greenhouse simulator, one with sample dropout ($p=0.8$, the choice of parameter $p$ will be analyzed in section \ref{sec:hyperparameter}) and one without sample dropout ($p=1$). Additionally, we adopt the soft actor-critic (SAC) algorithm, a widely used model-free RL algorithm, as a baseline for comparison. With $horizon = 120$, these algorithms are trained with a 120-day $Net profit$ as an optimization target.

As shown in the left of Fig. \ref{fig:results}, the algorithm with dropout converges better and has less variance than the one without dropout. This is mainly because the sample dropout module will discard a portion of samples with excessive feedback values, avoiding the local optimum. Meanwhile, the agent pays more attention to the worst-case states so that its variance is smaller. As for the SAC algorithm, it performs worse than our algorithm, which is caused by the low sample efficiency of the model-free method, making it difficult to learn enough information with limited samples.

Further, we evaluate the planting policies learned by the different algorithms on the tomato simulator, as shown in the right of Fig.~\ref{fig:results}. We observe that: (1) the policies have similar performance in the early stage when the crop is not growing; (2) when the crop starts to harvest, our algorithm outperforms both the SAC algorithm and skilled labor.

\subsection{Analysis of Robustness and Safety}
\label{sec:robustness}

\begin{figure}
\centering
\includegraphics[width=0.7\textwidth]{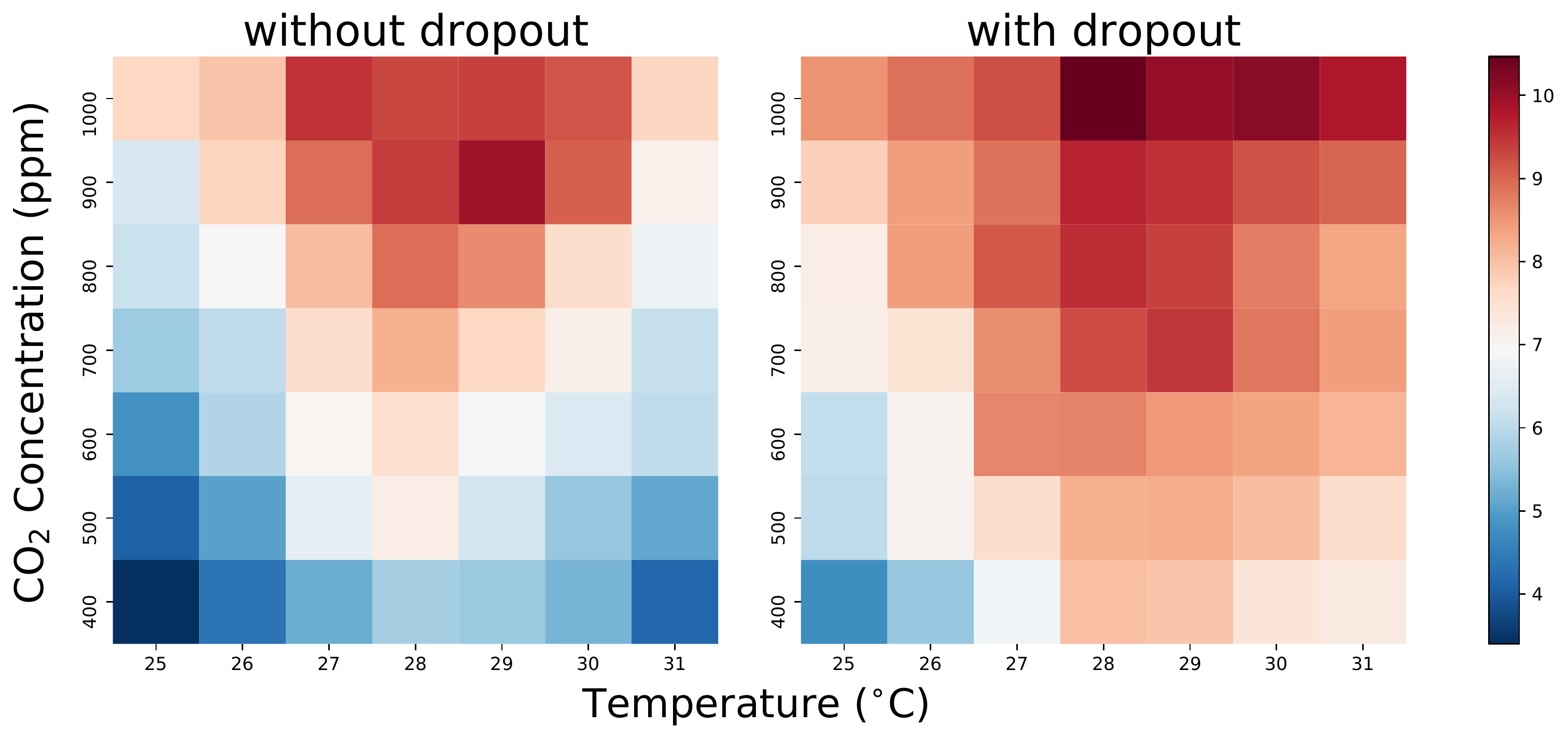}
\caption{The robustness performance is depicted as heat maps for various environment settings. The algorithms without ($p=1.0$) and with ($p=0.8$) dropout are conducted separately. Each pixel in the heat map represents the average reward in one specific experiment. The closer the color to red (hotter) means the higher reward, the better the algorithm performance in that environment, and vice versa. Each experiment stops after 500,000 steps.}
\label{fig:robustness}
\end{figure}

In order to verify the robustness improvement from the sample dropout module in our algorithm, we design a set of anti-disturbance experiments by perturbing the temperature ($^\circ$C) in the interval $[25, 31]$ and the $CO_2$ concentration (ppm) in the interval $[400, 1000]$. We test the algorithm with and without dropout separately (shown in Fig. \ref{fig:robustness}). 
In the heat map, each pixel represents the expected reward of the algorithm after training the same number of steps in each perturbed environment. Moreover, the closer the color to red (hotter) means a higher reward, and vice versa.
Obviously, the algorithm without dropout can only achieve a normally expected reward in a small area closer to the standard area. In contrast, the algorithm with dropout can maintain a higher expected reward in the more disturbed area, demonstrating that the sample dropout module can improve the robustness of the algorithm.

To further analyze the benefits of dropout, we set outside solar radiation (Iglob), greenhouse air temperature (AirT), and greenhouse air humidity (AirRH) as anomalous parameters, which are the critical factors to crop growth. 
According to these anomalous parameters, we test our algorithms in the simulator. Additionally, we use the fresh weight and retention rate of crops as indicators to evaluate the algorithm performance. The setting of anomalous parameters and the experimental results are shown in Table \ref{tab:exception-test}.

% We set up these abnormal conditions, train our algorithm again, and test on the black box growing environment of tomatoes. The results are shown in Table \ref{tab:exception-test}.

\begin{table}[ht]
\centering
\caption{Exception Test}
\label{tab:exception-test}
\begin{tabular}{c|cc|cc}
\toprule[1pt]
\multirow{2}{*}{\textbf{Parameter}}                                                 & \multicolumn{2}{c|}{\textbf{Fresh weight}}        & \multicolumn{2}{c}{\textbf{Retention rate}} \\ \cline{2-5} 
                                                                                    & $p=1$ & $p=0.8$ & $p=1$ & $p=0.8$ \\
\hline
\begin{tabular}[c]{@{}c@{}}$\text{AirT}\in(35,40)$\\\end{tabular} & 38.51                & \textbf{45.24}   & 80.23\%              & \textbf{85.35\%} \\ \hline
\begin{tabular}[c]{@{}c@{}}$\text{AirT}\in(-2,10)$\\ \end{tabular} & 30.31                & \textbf{38.49}   & 63.14\%              & \textbf{72.62\%} \\ \hline
\begin{tabular}[c]{@{}c@{}}$\text{AirRH}=90$\\ \end{tabular}           & 32.69                & \textbf{39.30}   & 68.10\%              & \textbf{74.16\%} \\ \hline
\begin{tabular}[c]{@{}c@{}}$\text{Iglob}=0$\\ \end{tabular}        & 36.76                & \textbf{43.71}   & 76.59\%              & \textbf{82.47\%} \\
\bottomrule[1pt]
\end{tabular}
\end{table}

Based on the results in Table~\ref{tab:exception-test}, we find that the algorithm with dropout has a higher fresh weight and retention rate under anomalous conditions, which shows higher safety, thus more promising for real-world applications.

% According to the results in the table above, it can be seen that under these anomalous conditions, although crops were affected and yield losses occurred, our method with dropout can retain a higher percentage of yield, confirming the conclusion that sample dropout improves the safety of the policy trained by the reinforcement learning algorithm.

\subsection{Analysis of Hyperparameter}
\label{sec:hyperparameter}

In this section, we investigate the sensitivity of our algorithm to the hyperparameter $p$ (details in Equation~\ref{def:batch-alpha-beta}). We vary the parameter $p$ from 1.0 to 0.6, representing the rate of discarded samples. Further downward adjustment of the parameter $p$ is no longer worth investigating, which is difficult for the algorithm to obtain enough information. This conclusion is confirmed in the following experiments.

\begin{figure}
\centering
\includegraphics[width=0.7\textwidth]{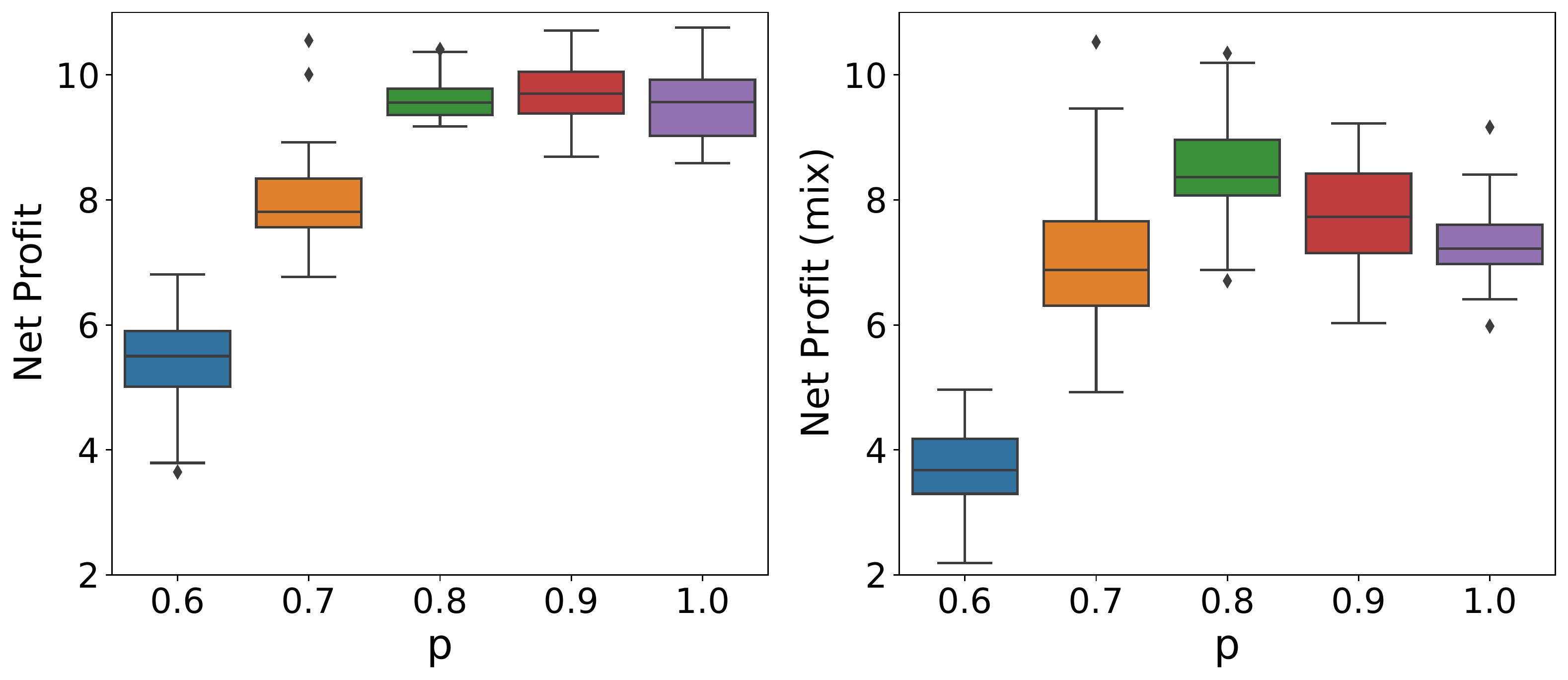}
\caption{The effect of adjusting parameter $p$: The left box plot shows the $Net profit$ that the algorithms can achieve with different parameters in the standard environment; The right box plot shows the average values of $Net profit$ that the algorithm can achieve with different parameters in the four disturbed environments (Temperature $\in \{26,30\}$; $CO_2$ concentration $\in \{500, 900\}$). Each experiment stops after 500,000 steps. The $x$-axis represents the parameter $p$ of dropout, and the $y$-axis represents the corresponding net profits.}
\label{fig:hyper-analysis}
\end{figure}

Firstly, We test the algorithm with dropout with different $p$ values ($p=1.0$ for no dropout) in the standard environment for multiple sets of experiments. The results are shown in the left of Fig. \ref{fig:hyper-analysis}. We observe that when $p \in \{0.8,0.9,1.0\}$, the corresponding $Net profit$ are close, implying that the algorithms have similar performance under these parameters. When $p < 0.8$, it can be seen that the performance of the algorithm decreases significantly. 

Next, we test the algorithm under different $p$ values with different disturbed environments. Then we take the mean value of these results as the final result. A larger mean value means better robustness of the algorithm. Specifically, we set up four different disturbed environments, controlling the temperature $\in \{26,30\}$ and the $CO_2$ concentration $\in \{500, 900\}$. The results are shown in the right of Fig. \ref{fig:hyper-analysis}. We observe that the optimal parameter value is 0.8. Moreover, a similarly significant decrease in the robustness of the algorithm starts when $p<0.8$.

% Based on the above experimental analysis, we find that our algorithm can get the best optimization with both good performance and robustness when $p=0.8$.

\section{Conclusions and Future Work}
%Robust Model-based Reinforcement Learning for Autonomous
In this work, we propose a robust model-based RL framework to alleviate the sample inefficiency and safety concern in greenhouse automation. To be specific, our framework utilizes sensors and actuators deployed in the greenhouse to collect observation and action samples to learn an ensemble of environment models and optimize the policy in a Dyna-style manner. Experimental results demonstrate the effectiveness and superiority of our framework in terms of robustness and efficiency, contribute to better crop growth with a higher safety guarantee.

% We have experimentally demonstrated that our approach can learn more valuable planting policies than traditional Genetic Algorithms and state-of-the-art RL algorithms like SAC. Besides, through sample dropout, our approach can effectively improve the planting policy's stability and safety and still maintain better performance in the face of extreme and abnormal conditions.

Our future work will incorporate more prior knowledge of agriculture to improve the simulator. We will also deploy the planting policies trained by the real greenhouses algorithm to evaluate the framework through long-term real-world experiments further. Besides, we plan to use offline RL methods to improve sample utilization, reduce training costs, and use meta RL methods to transfer learning different crop species to improve the algorithm's generalization performance.

\section*{Acknowledgments}
This work was done when Wanpeng Zhang, Xiaoyan Cao, Yao Yao and Zhicheng An was intern at Tencent AI Lab. This work was also supported in part by the National Natural Science Foundation of China (61972219), the Research and Development Program of Shenzhen (JCYJ20190813174403598, SGDX20190918101201696), the National Key Research and Development Program of China (2018YFB1800601), and the Overseas Research Cooperation Fund of Tsinghua Shenzhen International Graduate School
(HW2021013).

\bibliography{acml21}

\end{document}